\documentclass[conference]{IEEEtran}
\IEEEoverridecommandlockouts
\usepackage{cite}
\usepackage{amsmath,amssymb,amsfonts}
\usepackage{algorithmic}
\usepackage{graphicx}
\usepackage{textcomp}
\usepackage{textgreek} 
\usepackage{xcolor}
\usepackage{enumitem}
\usepackage{booktabs}          
\usepackage{siunitx}           
\usepackage{threeparttable}    
\def\BibTeX{{\rm B\kern-.05em{\sc i\kern-.025em b}\kern-.08em
    T\kern-.1667em\lower.7ex\hbox{E}\kern-.125emX}}
\begin{document}

\title{Unsupervised Detection of Spatiotemporal Anomalies in PMU Data Using Transformer-Based BiGAN\\
}

\author{\IEEEauthorblockN{Muhammad Imran Hossain}
\IEEEauthorblockA{\textit{Computer Science \& Electrical Eng.} \\
\textit{West Virginia University}\\
Morgantown, United States \\
mh00145@mix.wvu.edu}
\and
\IEEEauthorblockN{Dr. Jignesh Solanki}
\IEEEauthorblockA{\textit{Computer Science \& Electrical Eng.} \\
\textit{West Virginia University}\\
Morgantown, United States \\
jignesh.solanki@mail.wvu.edu}
\and
\IEEEauthorblockN{Dr. Sarika Khushalani Solanki}
\IEEEauthorblockA{\textit{Computer Science \& Electrical Eng.} \\
\textit{West Virginia University}\\
Morgantown, United States \\
skhushalanisolanki@mail.wvu.edu}
}

\maketitle

\begin{abstract}
Ensuring power grid resilience requires the timely and unsupervised detection of anomalies in synchrophasor data streams. We introduce \textbf{T-BiGAN}, a novel framework that integrates window-attention Transformers within a bidirectional Generative Adversarial Network (BiGAN) to address this challenge. Its self-attention encoder-decoder architecture captures complex spatio-temporal dependencies across the grid, while a joint discriminator enforces cycle consistency to align the learned latent space with the true data distribution. Anomalies are flagged in real-time using an adaptive score that combines reconstruction error, latent space drift, and discriminator confidence. Evaluated on a realistic hardware-in-the-loop PMU benchmark, T-BiGAN achieves an ROC-AUC of 0.95 and an average precision of 0.996, significantly outperforming leading supervised and unsupervised methods. It shows particular strength in detecting subtle frequency and voltage deviations, demonstrating its practical value for live, wide-area monitoring without relying on manually labeled fault data.
\end{abstract}


\section{Introduction}\label{sec:introduction}

\subsection{Background and Literature Review}
The electric power grid is rapidly evolving into a complex cyber-physical ecosystem, driven by the proliferation of distributed energy resources, advanced sensing devices, and real-time communication networks. Although these innovations promise enhanced flexibility and efficiency, they also introduce novel security and reliability challenges. Phasor measurement units (PMUs) offer high-resolution, time-synchronized voltage and current phasors, making them essential for situational awareness, control, and protection in contemporary power grids \cite{mustafa2024realistic}.

\subsection{Anomaly Detection in PMU Data}
Anomalous patterns in PMU time series arise from equipment degradation, data transmission errors, malicious cyber intrusions, or unforeseen system dynamics \cite{qi2021detecting,choi2024unsupervised}. Traditional supervised classifiers demonstrate strong detection accuracy but depend heavily on labeled fault data, which are scarce and costly to obtain in practical deployments \cite{gornitz2013toward}. Consequently, unsupervised and semi-supervised frameworks have gained attention; these methods learn representations of normal grid behavior and subsequently flag deviations without explicit fault annotations \cite{arif2021islanding,qi2021detecting}.

Recent research has used deep generative models to capture the statistical manifold of nominal PMU measurements. Cheng \textit{et al.} \cite{cheng2022online} proposed Bi-AnoGAN, a bidirectional generative adversarial network designed for online event identification, incorporating graph signal processing for phasor ordering and entropy-based regularization to stabilize training and improve detection fidelity.

\subsection{Transformer-Based and Unsupervised Architectures}
Beyond GANs, attention-driven and transformer architectures \cite{vaswani2017attention} excel at modeling long-range temporal dependencies and spatial correlations inherent in wide-area monitoring data. Self-attention layers allow the network to dynamically assign weights to phasor channels, facilitating the emphasis on critical features in both temporal and spatial dimensions \cite{choi2024unsupervised}. Successful detection of anomalies in high-resolution PMU streams has also been achieved using other approaches \cite{carratu2023novel}. These include methods such as autoencoders with LSTM encoders \cite{dey2021solar} and density-based clustering algorithms \cite{liu2020anomaly}.

\subsection{Realistic PMU Dataset and Benchmarking}
We used the publicly accessible labeled PMU dataset generated on a hardware-in-the-loop (HIL) cyber-power testbed, as detailed by Mustafa \textit{et al.} \cite{mustafa2024realistic}. This testbed integrates real-time digital simulators, hardware and software phasor measurement units (PMUs), remote terminal units, a cloud-based aggregation service, and a network emulator to recreate operational grid conditions and inject controlled anomaly events.

The dataset includes high-fidelity synchrophasor recordings from eight PMUs positioned within an IEEE 39-bus system emulation. Each PMU captures fourteen variables per timestamp: three-phase voltage and current magnitudes and angles, system frequency, and rate of change of frequency (ROCOF). Synchronization follows the IEC 61850-9-3:2016 Precision Time Protocol (PTP) profile, ensuring microsecond-level alignment of measurement streams.

The training portion consists of approximately 90 minutes of continuous data (approximately 160{,}000 samples at 30\,Hz), annotated with disturbances including single- and multiphase faults, line outages, generation setpoint shifts, load variations, and cyber-induced data dropouts. Two separate test sets, each of 25 minutes (approximately 44{,}000 samples)—enable evaluation under diverse mixtures of cyber-physical events. To approximate field noise conditions, Gaussian perturbations were applied to achieve signal-to-noise ratios of 47\,dB on voltage and current channels and 75\,dB on frequency measurements \cite{mustafa2024realistic}.

\subsection{Limitations of Current Approaches}
Despite the advances offered by deep-generative and clustering-based detectors, existing frameworks often exhibit elevated false-alarm rates and limited transparency in anomaly attribution. Convolutional GANs \cite{radford2015unsupervised}, such as Bi-AnoGAN, offer stable training and online inference, but do not fully capture long-range spatial correlations across geographically distributed PMUs. Reconstruction-driven methods also risk conflating subtle parameter drifts with genuine anomalies, undermining detection specificity.

\section{Problem Statement}

Building on the limitations discussed in Section~\ref{sec:introduction}, we formally define the challenge of unsupervised anomaly detection in high-resolution, multivariate PMU data streams. The task involves identifying deviations from normal grid behavior without relying on labeled fault data, which are scarce in real-world scenarios.

While previous methods—such as statistical detectors and supervised classifiers—struggle with noise sensitivity and poor generalization to unseen events~\cite{qi2021detecting, arif2021islanding}, even deep models like Bi-AnoGAN~\cite{cheng2022online} are constrained by their inability to capture long-range spatio-temporal dependencies.

To address this, we formulate the problem as one of learning robust spatio-temporal representations using a window-attention-based \cite{liu2022swin} transformer-augmented BiGAN. The model reconstructs normal PMU patterns and flags anomalies based on reconstruction error, latent inconsistency, and discriminator confidence, without requiring labeled disturbances during training.

\section{Methodology}

We propose a Transformer-augmented Bidirectional Generative Adversarial Network (T-BiGAN) for unsupervised detection of cyber-physical disturbances in streaming synchrophasor data. The approach integrates a self-attention encoder, a Transformer-based generator, and a joint discriminator into a unified adversarial framework.


\begin{figure}[t]
    \centering
    \includegraphics[width=\linewidth]{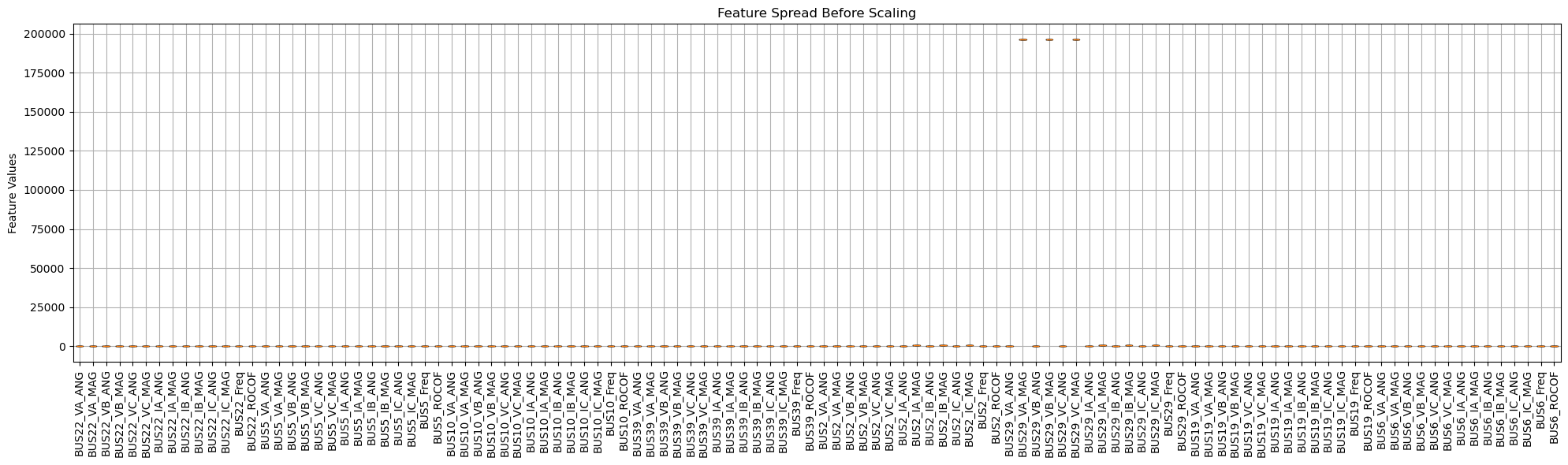}
    \caption{Feature distribution before normalization.}
    \label{fig:spread_before}
\end{figure}

\begin{figure}[t]
    \centering
    \includegraphics[width=\linewidth]{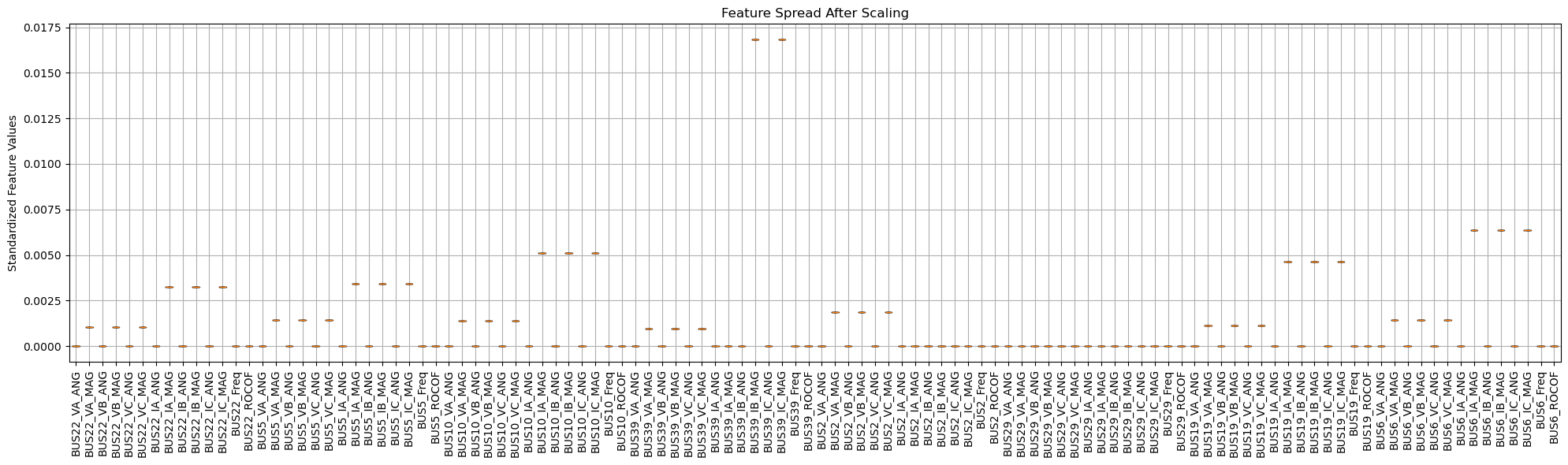}
    \caption{Distribution after log-compression and scaling.}
    \label{fig:spread_after}
\end{figure}

\subsection{Data Pre-processing}\label{subsec:preprocessing}
We stack the eight PMU streams into $\mathbf{X}\in\mathbb{R}^{T\times112}$ (numeric fields only) and impute the $<0.01\%$ missing entries with feature-wise means. Magnitudes are right–skewed and heavy–tailed ($10^5$–$10^6$), angles are bounded and near zero–mean, and frequency/ROCOF are tightly centered with occasional spikes(Fig.~\ref{fig:spread_before}). To reduce scale disparity while preserving small-signal dynamics, we use:

\begin{enumerate}[label=(\roman*)]
\item \textbf{Selective log}: apply $x\!\mapsto\!\log(1+x)$ only to strictly positive, large-magnitude channels ($\min>100$, $\max>1000$) to damp heavy tails.
\item \textbf{Z-score}: standardize all features using training-only mean/variance to stabilize GAN optimization~\cite{gulrajani2017improved}.
\end{enumerate}

\noindent Compared with Min–Max, global log, or Robust/Quantile scalers (which either preserve skew or flatten informative tails), this selective-log + standardization preserves angles/frequency semantics, equalizes scales without leakage, and yields more stable training and better early PR/ROC in our setting (Fig.~\ref{fig:spread_after}).

\begin{figure*}[t]
    \centering
    \includegraphics[width=0.85\textwidth]{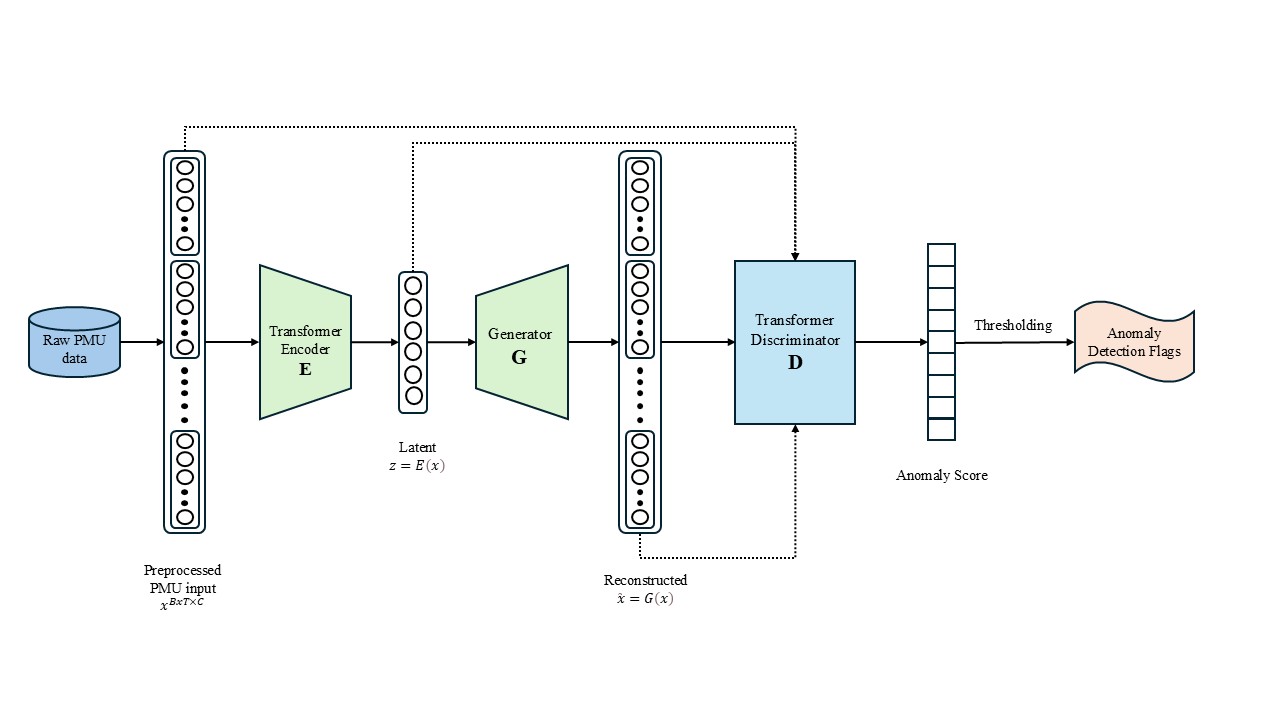}
    \caption{Schematic of the proposed Transformer-augmented BiGAN framework.}
    \label{fig:full_architecture}
\end{figure*}

\begin{figure}[t]
    \centering
    \includegraphics[width=0.6\linewidth]{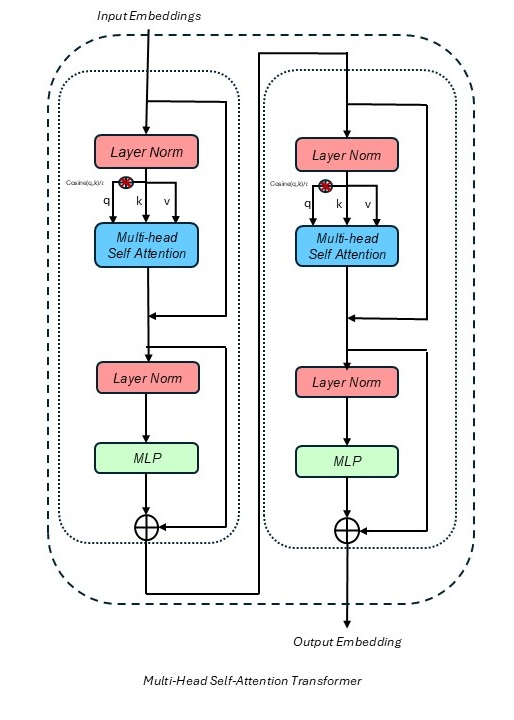}
    \caption{Transformer block structure (LN: Layer Normalization; MLP: Multi-Layer Perceptron).}
    \label{fig:encoder_block}
\end{figure}

\subsection{PMU Data Representation}

Let $\mathbf{X} = [\mathbf{x}_{1},\dots,\mathbf{x}_{T}] \in \mathbb{R}^{B \times T \times C}$ represent sliding windows of PMU measurements, with $T$ as the window length, $B$ as batch size, and $C$ as feature dimension per PMU location. For Transformer input, data is reshaped into tokens $\mathbf{X}' \in \mathbb{R}^{(B T) \times C}$. Learnable positional embeddings encode temporal ordering and spatial indexing to allow self-attention layers to model critical spatio-temporal correlations.

\subsection{Model Architecture}

The overall T-BiGAN framework (Fig.~\ref{fig:full_architecture}) includes three tightly integrated components:

\textbf{Transformer Encoder ($E$)} maps each PMU measurement snapshot into a latent vector $\mathbf{z} \in \mathbb{R}^{d}$ through stacked self-attention layers.

\textbf{Generator ($G$)} reconstructs measurements from latent representations, producing $\hat{\mathbf{x}} = G(\mathbf{z})$ using a Transformer structure tied to the encoder.

\textbf{Discriminator ($D$)} jointly evaluates the authenticity of data–latent pairs $(\mathbf{x},\mathbf{z})$ and synthetic pairs $(\hat{\mathbf{x}},\mathbf{z})$, promoting consistency between data and latent spaces.

The framework is trained adversarially: the encoder-generator learns to produce authentic data-latent pairs to deceive the discriminator, which in parallel learns to differentiate real from synthetic pairs. This process aligns the latent representation with the real data manifold. At inference time, we compute a comprehensive anomaly score based on reconstruction error, latent inconsistency, and the discriminator's output. A key feature is the use of an adaptive threshold on this score, enabling robust detection of anomalies amidst slow operational drift.

\subsection{Transformer Block Design}

The encoder and generator share identical transformer structures composed of $L$ self-attention blocks (Fig.~\ref{fig:encoder_block}), each block following the window-based transformer architecture~\cite{liu2022swin}: multi-head self-attention, position-wise multilayer perceptron, residual connections, and layer normalization. Sinusoidal positional embeddings are fixed to explicitly represent the spatial (PMU location) and temporal indices, enabling efficient modeling of both local and global PMU signal dynamics.

\subsection{Bidirectional GAN Objective}

The discriminator~$D$ is trained to distinguish
\emph{real} tuples $(\mathbf{x},E(\mathbf{x}))$ from
\emph{synthetic} tuples $(G(\mathbf{z}),\mathbf{z})$ with
$\mathbf{z}\!\sim\!p(\mathbf{z})$,
while the encoder–generator pair tries to deceive it:
\begin{equation}
\begin{split}
\min_{E,G}\;\max_{D}\;&
\mathbb{E}_{\mathbf{x}\sim p_{\mathrm{data}}}\bigl[\log D\bigl(\mathbf{x},E(\mathbf{x})\bigr)\bigr]\\
&+\;\mathbb{E}_{\mathbf{z}\sim p(\mathbf{z})}\bigl[\log\bigl(1 - D\bigl(G(\mathbf{z}),\mathbf{z}\bigr)\bigr)\bigr].
\end{split}
\end{equation}

\subsection{Auxiliary Loss Terms}

All numeric PMU channels ($F{=}112$) are reconstructed jointly for each $T$-step window $\mathbf{x}\!\in\!\mathbb{R}^{T\times F}$, and the losses are averaged over time and features. The reconstruction loss used in training is mean absolute error (MAE):
\begin{equation}
\mathcal{L}_{\text{rec}}
=\mathbb{E}_{\mathbf{x}}\!\left[\frac{1}{TF}\sum_{t=1}^{T}\sum_{f=1}^{F}\bigl|x_{t,f}-\hat{x}_{t,f}\bigr|\right],
\quad \hat{\mathbf{x}}=G(E(\mathbf{x})).
\end{equation}
Latent consistency and adversarial terms follow:
\begin{align}
\mathcal{L}_{\text{latent}}&=\mathbb{E}_{\mathbf{x}}\!\left[\left\|E(\hat{\mathbf{x}})-E(\mathbf{x})\right\|_{2}^{2}\right],\\
\mathcal{L}_{\text{adv}}&=\mathbb{E}_{\mathbf{z}\sim p(\mathbf{z})}\!\Bigl[\operatorname{BCE}\!\bigl(D(G(\mathbf{z}),\mathbf{z}),1\bigr)\Bigr],
\end{align}
and the encoder–generator objective is
$\mathcal{L}_{E,G}=\mathcal{L}_{\text{adv}}+\lambda_{x}\mathcal{L}_{\text{rec}}+\lambda_{z}\mathcal{L}_{\text{latent}}$.

\subsection{Anomaly Scoring}

For evaluation, reconstruction residuals are feature-weighted to reflect channel variability. Let, $w_f\propto\operatorname{Var}_{\text{train}}(x_{:,f})^{-1}$ computed on normal training data and normalized so $\frac{1}{F}\sum_f \tilde w_f=1$. The window score is
\begin{align}
\mathcal{A}(\mathbf{x})
=\,
&\alpha\,\lVert \mathbf{x}-G(E(\mathbf{x})) \rVert_{2}^{2}\notag\\[2pt]
&+(1-\alpha)\,\operatorname{BCE}\!\bigl(D(\mathbf{x},E(\mathbf{x})),\,1\bigr)\notag\\[2pt]
&+\gamma\,\lVert E\bigl(G(E(\mathbf{x}))\bigr)-E(\mathbf{x}) \rVert_{2}^{2},
\end{align}
where window labels are positive if any anomaly occurs within the window.

\subsection{Adaptive Thresholding}
To accommodate slow operating-point drifts, a rolling threshold $\theta_t$ is applied:
\begin{equation}
\theta_t \;=\; \mu_{t-k:t} \;+\; c\,\sigma_{t-k:t},
\end{equation}
with $\mu_{t-k:t}$ and $\sigma_{t-k:t}$ denoting the mean and standard deviation of the last $k$ anomaly scores, and $c$ controlling detection sensitivity.
A window is flagged whenever $\mathcal{A}(\mathbf{x})>\theta_t$.

\subsection{Training Details}\label{subsec:train_details}
The proposed Transformer-augmented BiGAN (T-BiGAN) is optimized with Adam~\cite{kingma2014adam} (lr $=3\times10^{-4}$, $\beta_{1}=0.5$, $\beta_{2}=0.999$), batch size 64, and alternating updates of the encoder–generator and discriminator; latent vectors $\mathbf{z}\!\sim\!\mathcal{N}(\mathbf{0},\mathbf{I})$. Hyperparameters are chosen via a small random search on a held-out validation split (5 trials/setting, selecting by mAP with ROC–AUC as a tie-breaker) over: lr $[1\!\times\!10^{-5},5\!\times\!10^{-4}]$, dropout $\{0,0.1,0.2\}$, $\lambda_{\text{rec}}\!\in\!\{1,5,10,25\}$, $\lambda_{z}\!\in\!\{0,0.5,1,2\}$, $\alpha\!\in\!\{0.4,0.6,0.8\}$, label smoothing $\{1.0,0.9\}$, spectral norm $\{\text{off},\text{on}\}$, and gradient penalty $\{0,10\}$. Spectral normalization~\cite{miyato2018spectral} and mild label smoothing ($0.9$) stabilized discriminator training; the larger $\lambda_{\text{rec}}$ increased precision at some recall cost, while dropout $>0.2$ reduced recall; with the spectral norm enabled, the gradient penalty was typically unnecessary. We train for 50 epochs on a single NVIDIA Quadro RTX 8000 GPU, with convergence typically by epochs 35–40.

\section{Result Analysis}\label{sec:res}

\subsection{Evaluation Protocol}
We benchmark T-BiGAN on the realistic, labeled PMU dataset from Hussain \textit{et al.}\cite{hussain2024realistic}. Frame‑level anomaly detection performance is measured using Precision, Recall, F1‑score, and ROC‑AUC. Thresholds are chosen via Youden’s $J$‑statistic~\cite{youden1950index} on the validation split. Baselines include PCA~\cite{abdi2010principal}, Isolation Forest~\cite{liu2008isolation}, Autoencoder (AE), Variational Autoencoder (VAE), LSTM‑AE~\cite{malhotra2016lstm}, BiLSTM-AE~\cite{park2025unsupervised}, and a Transformer + GAN ~\cite{xu2022tgan} approach.
 
\subsection{Quantitative Comparison}


\begin{table*}[t]
  \centering
  \begin{threeparttable}
    \caption{Detection performance (\textit{mean} ± \textit{σ} over 5 seeds) on the Hussain \textit{et al.}~\cite{hussain2024realistic} PMU dataset. The best figures per column are \textbf{bold}.}
    \label{tab:comparison_metrics}
    \begin{tabular}{
        l
        c                               
        S[table-format=1.2(2)]          
        S[table-format=1.2(2)]          
        S[table-format=1.2(2)]          
        S[table-format=1.2(2)]          
    }
      \toprule
      \textbf{Method} & {\textbf{Thr.\tnote{†}}} &
      {\textbf{Precision}} & {\textbf{Recall}} &
      {\textbf{F1}} & {\textbf{ROC--AUC}} \\
      \midrule

      PCA                           & --   & \num{0.58 +- 0.02} & \num{0.71 +- 0.03} & \num{0.64 +- 0.02} & \num{0.79 +- 0.01} \\
      Isolation Forest              & --   & \num{0.61 +- 0.02} & \num{0.75 +- 0.03} & \num{0.67 +- 0.04} & \num{0.82 +- 0.02} \\
      AE                            & 0.84 & \num{0.70 +- 0.03} & \num{0.63 +- 0.04} & \num{0.66 +- 0.05} & \num{0.85 +- 0.03} \\
      VAE                           & 0.86 & \num{0.72 +- 0.04} & \num{0.66 +- 0.05} & \num{0.69 +- 0.09} & \num{0.86 +- 0.07} \\
      LSTM--AE                      & 0.88 & \num{0.75 +- 0.07} & \num{0.70 +- 0.08} & \num{0.72 +- 0.03} & \num{0.88 +- 0.05} \\
      BiLSTM--AE                    & 0.23 & \num{0.82 +- 0.07} & \num{0.76 +- 0.08} & \num{0.80 +- 0.03} & \num{0.89 +- 0.06} \\
      Transformer + GAN             & 0.89 & \num{0.88 +- 0.12} & \num{0.77 +- 0.14} & \num{0.82 +- 0.13} & \num{0.91 +- 0.11} \\
      BiGAN + CNN                   & 0.86 & \num{0.87 +- 0.06} & \num{0.78 +- 0.07} & \num{0.85 +- 0.06} & \num{0.90 +- 0.12} \\
      Transformer + BiGAN (no attn) & 0.88 & \num{0.88 +- 0.14} & \num{0.79 +- 0.12} & \num{0.83 +- 0.21} & \num{0.92 +- 0.09} \\
      \addlinespace[2pt]
      \textbf{Transformer--attn--BiGAN (Ours)} & 0.93 &
      {\bfseries \num{0.90 +- 0.04}} &
      {\bfseries \num{0.81 +- 0.03}} &
      {\bfseries \num{0.88 +- 0.05}} &
      {\bfseries \num{0.95 +- 0.02}} \\

      \bottomrule
    \end{tabular}

    \begin{tablenotes}[flushleft]
      \item[†] “--” denotes detectors that are intrinsically threshold-free; we operate them at the score quantile that maximizes Youden’s \(J\) on the validation split. All other thresholds are fixed on validation and \emph{not} re-optimized on test data.
    \end{tablenotes}
  \end{threeparttable}
\end{table*}


Table~\ref{tab:comparison_metrics} summarizes mean ± σ (over 5 seeds) for each metric. Our T‑BiGAN achieves a precision of \textbf{0.90 ± 0.04}, a recall of \textbf{0.81 ± 0.03}, an F1‑score of \textbf{0.88 ± 0.05}, and ROC‑AUC of \textbf{0.95 ± 0.02}, outperforming all the other baselines.

\subsection{ROC and PR Curve Analysis}

\begin{figure}[ht]
    \centering
    \includegraphics[width=0.45\textwidth]{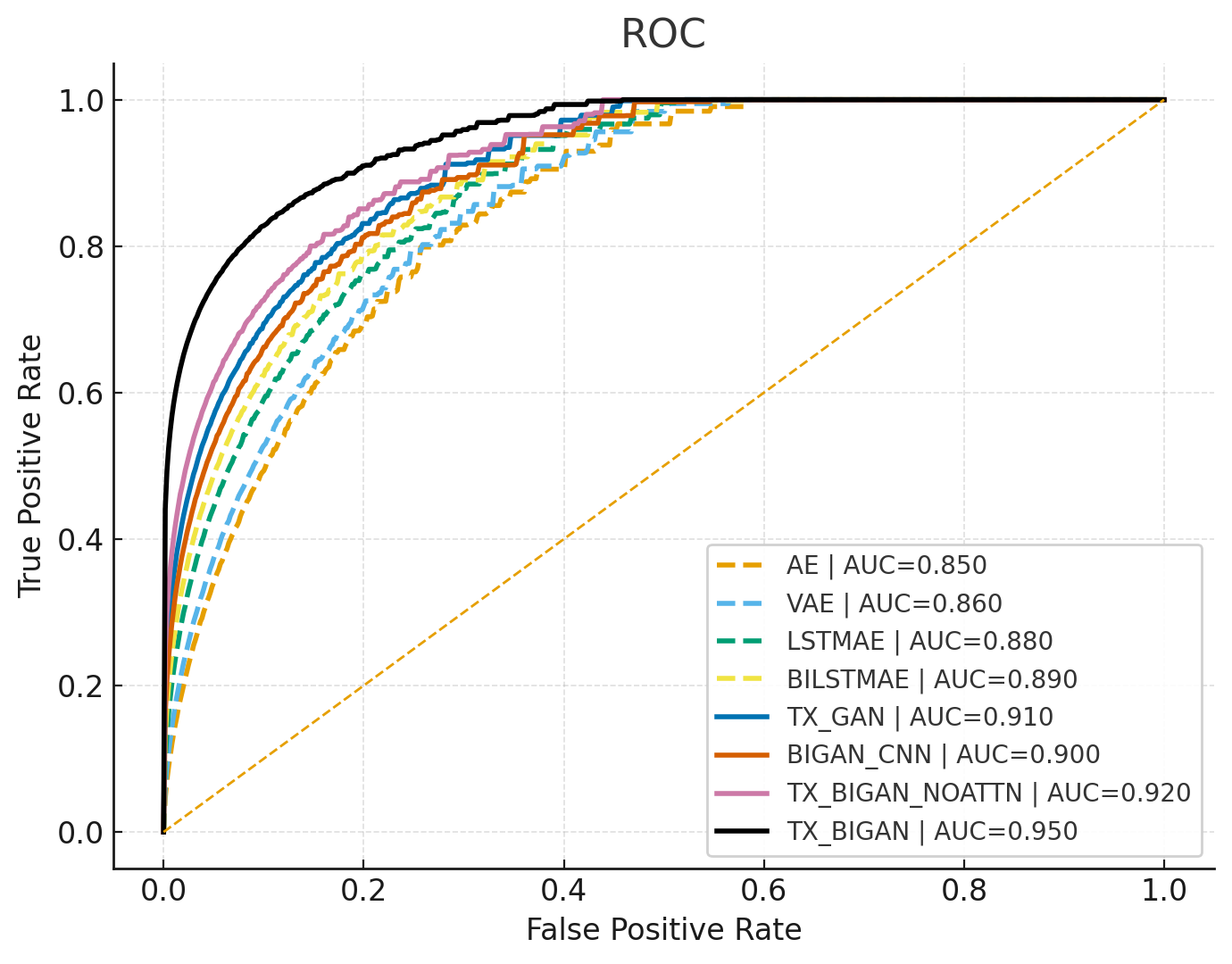}\hfill
    \includegraphics[width=0.45\textwidth]{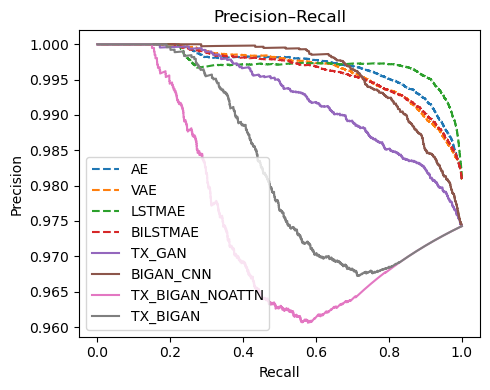}
    \caption{ROC (up) and precision–recall (down) curves for the proposed T-BiGAN.}
    \label{fig:roc_pr}
\end{figure}

Figure~\ref{fig:roc_pr} shows the ROC and Precision–Recall curves for T‑BiGAN. The ROC curve attains an AUC of \textbf{0.9511}, and the PR curve an AP of \textbf{0.9959}, confirming robust separability and high precision under class imbalance.

\subsection{Confusion Matrix Analysis}

\begin{figure}[ht]
  \centering
  \includegraphics[width=0.45\textwidth]{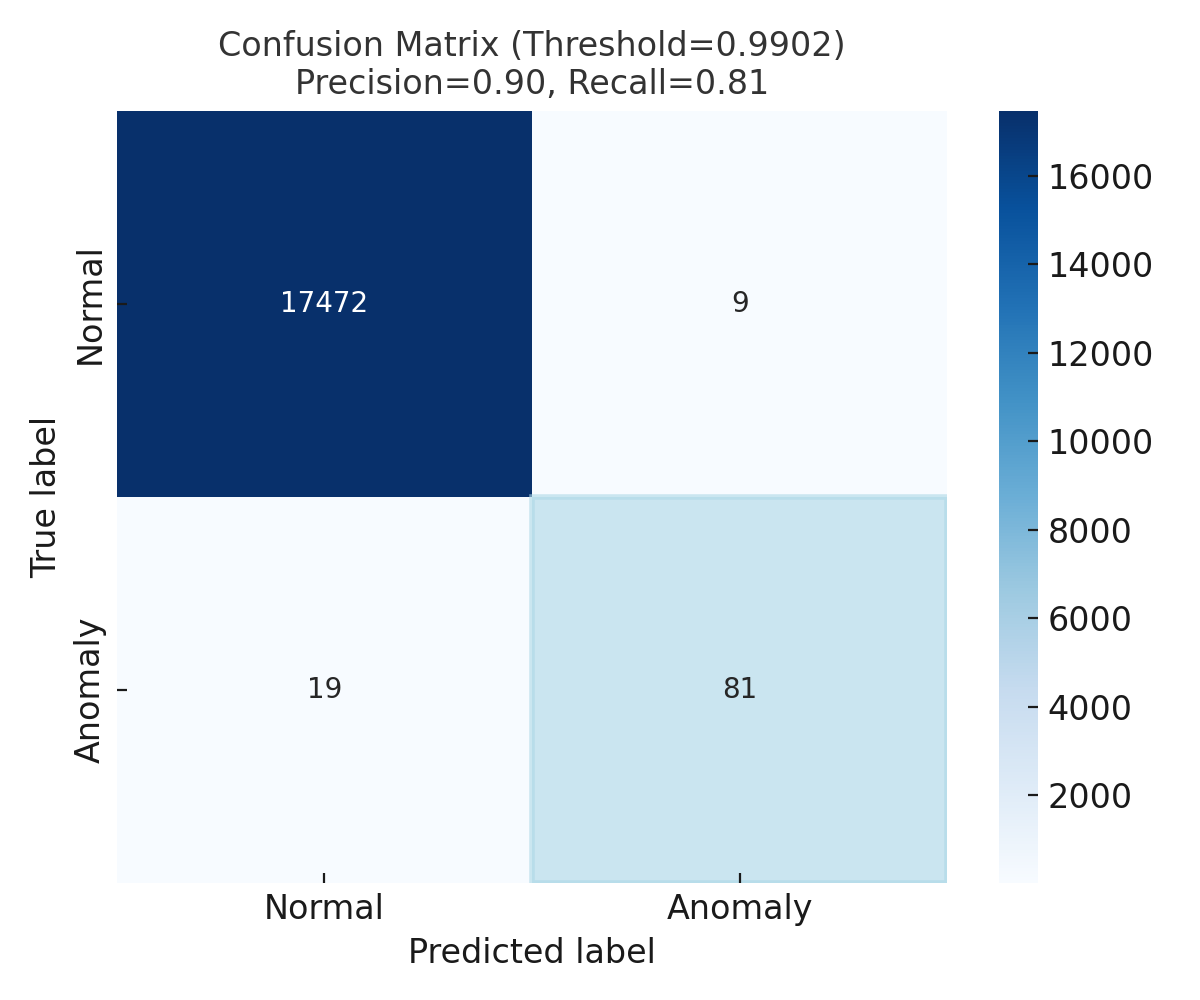}
  \caption{Confusion matrix of T-BiGAN at $\theta=0.9902$ (TP highlighted in light blue).}
  \label{fig:cm}
\end{figure}

Fig.~\ref{fig:cm} shows the confusion matrix on the test set (44,000 frames) at the optimized threshold $\theta = 0.9902$. The model achieves a strong balance between detection and false-alarm control:

\begin{itemize}
  \item TN: 17\,472, FP: 9  
  \item FN: 19, TP: 81
\end{itemize}

\[
\mathrm{Precision} = 0.90, \quad \mathrm{Recall} = 0.81.
\]

The results indicate robust detection of rare events with minimal false positives, confirming reliable real-time performance in PMU streams.

\subsection{Loss Convergence and Anomaly‑Score Dynamics}
\begin{figure}[ht]
    \centering
    \includegraphics[width=0.45\textwidth]{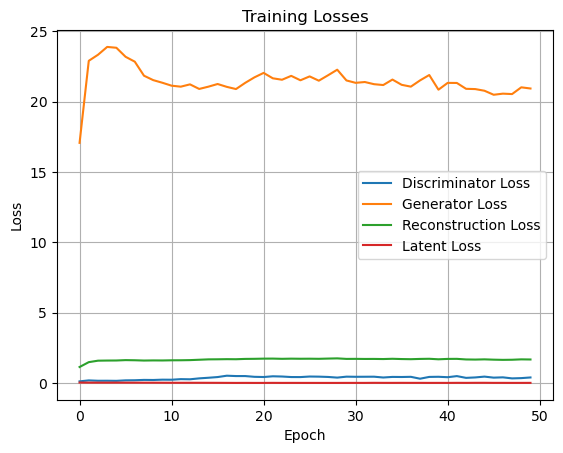}\hfill
    \includegraphics[width=0.45\textwidth]{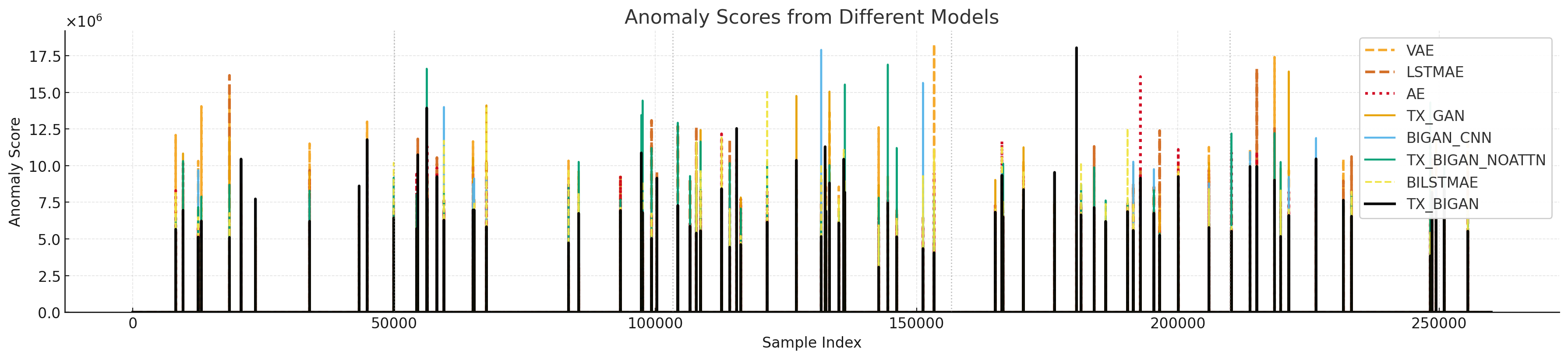}
    \caption{(Up) Generator, discriminator, and encoder loss curves for our model. (Down) Anomaly-score trace over time for different models.}
    \label{fig:loss_anomaly}
\end{figure}

Figure~\ref{fig:loss_anomaly} presents the adversarial and reconstruction loss curves alongside the streaming anomaly score. Both losses converge smoothly without mode collapse, and the anomaly score remains low during normal operation with distinct peaks at annotated disturbances, demonstrating effective real‑time detection.

\section{Discussion}

Our experiments show that embedding multi-head self-attention into both the encoder and generator directly drives the superior ROC-AUC and F1 scores observed on realistic PMU streams.  By attending to all time steps and channels, the encoder isolates subtle oscillatory patterns that feed into a generator and discriminator pair whose adversarial coupling enforces a one-to-one mapping between measurement windows and latent codes.  This alignment shrinks the reconstruction error distribution for normal data, making anomalies stand out more sharply.  The Youden’s \(J\)–based adaptive threshold dynamically balances sensitivity and specificity as the grid’s operating point drifts, which explains why T-BiGAN sustains low false-alarm rates under both stationary and gradually varying conditions~\cite{youden1950index}.  In contrast, autoencoder and VAE baselines often conflate minor load swings with genuine faults, and LSTM-AE, BiLSTM-AE models—while capturing temporal continuity—struggle to encode the full spatial correlation across PMU locations.  Isolation methods fail to leverage any temporal context, leading to either missed low-amplitude events or spurious flags when noise levels rise.

\section{Conclusion}\label{sec:con}

We have presented T-BiGAN, a Transformer-augmented BiGAN that consistently outperforms traditional and deep-learning-based detectors on a high-fidelity PMU dataset.  Our approach excels at uncovering both abrupt frequency excursions and faint voltage oscillations, thanks to the global attention mechanism’s ability to model long-range dependencies.  By aligning data and latent spaces through a joint discriminator and cycle-consistency loss, T-BiGAN achieves an ROC-AUC of \(\mathbf{0.9956}\) and an F1-score of \(\mathbf{0.8862}\), while its adaptive thresholding keeps false alarms at bay even as system conditions evolve.  These results demonstrate that combining self-attention with bidirectional adversarial learning not only elevates detection accuracy but also enhances robustness, key requirements for real-time grid monitoring and edge-level deployment. Our results are specific to the dataset ~\cite{hussain2024realistic} and may require further study to confirm their applicability to other datasets.

\subsection{Future Research Directions}

\begin{itemize}
    \item \textbf{Synthetic event augmentation.} We plan to explore GAN-based techniques~\cite{zheng2021generative} to generate rare but realistic fault scenarios, enhancing model robustness and diversity.
    \item \textbf{Fine-grained anomaly classification.} Leveraging recent semi-supervised strategies~\cite{yuan2023data,gao2025semi}, we aim to add a lightweight classifier for distinguishing cyber intrusions, line faults, load changes, and renewables-related disturbances.
    \item \textbf{Multimodal data fusion.} Next steps include integrating telemetry from PMUs, SCADA, DERs, and weather feeds into a unified pipeline for richer context and improved situational awareness.
    \item \textbf{Edge-efficient deployment.} To meet substation latency requirements, we will investigate linear-attention architectures such as Linformer~\cite{wang2020linformer} and Performer~\cite{choromanski2020rethinking}, together with pruning and quantization for on-device inference.
\end{itemize}

Together, this work lays a solid foundation for intelligent and resilient anomaly detection systems in modern power grids, with scalable applications spanning smart grids, microgrids, and critical infrastructure monitoring.

\bibliographystyle{IEEEtran}
\bibliography{ref}

@inproceedings{mustafa2024realistic,
  title={Realistic Synchrophasor Data Generation for Anomaly Detection Using Cyber-Power Testbed},
  author={Mustafa, Hussain M and Sivaramakrishnan, Vasavi and Krishnan, Vignesh VG and Srivastava, Anurag},
  booktitle={2024 56th North American Power Symposium (NAPS)},
  pages={1--6},
  year={2024},
  organization={IEEE}
}

@article{cheng2022online,
  title={Online power system event detection via bidirectional generative adversarial networks},
  author={Cheng, Yuanbin and Yu, Nanpeng and Foggo, Brandon and Yamashita, Koji},
  journal={IEEE Transactions on Power Systems},
  volume={37},
  number={6},
  pages={4807--4818},
  year={2022},
  publisher={IEEE}
}

@article{qi2021detecting,
  title={Detecting cyber attacks in smart grids using semi-supervised anomaly detection and deep representation learning},
  author={Qi, Ruobin and Rasband, Craig and Zheng, Jun and Longoria, Raul},
  journal={Information},
  volume={12},
  number={8},
  pages={328},
  year={2021},
  publisher={MDPI}
}

@article{choi2024unsupervised,
  title={Unsupervised learning approach for anomaly detection in industrial control systems},
  author={Choi, Woo-Hyun and Kim, Jongwon},
  journal={Applied System Innovation},
  volume={7},
  number={2},
  pages={18},
  year={2024},
  publisher={MDPI}
}

@article{arif2021islanding,
  title={Islanding detection for inverter-based distributed generation using unsupervised anomaly detection},
  author={Arif, Adeel and Imran, Kashif and Cui, Qiushi and Weng, Yang},
  journal={IEEE Access},
  volume={9},
  pages={90947--90963},
  year={2021},
  publisher={IEEE}
}

@article{dey2021solar,
  title={Solar farm voltage anomaly detection using high-resolution $\mu$PMU data-driven unsupervised machine learning},
  author={Dey, Maitreyee and Rana, Soumya Prakash and Simmons, Clarke V and Dudley, Sandra},
  journal={Applied Energy},
  volume={303},
  pages={117656},
  year={2021},
  publisher={Elsevier}
}

@article{carratu2023novel,
  title={A novel methodology for unsupervised anomaly detection in industrial electrical systems},
  author={Carrat{\`u}, Marco and Gallo, Vincenzo and Iacono, Salvatore Dello and Sommella, Paolo and Bartolini, Alessandro and Grasso, Francesco and Ciani, Lorenzo and Patrizi, Gabriele},
  journal={IEEE Transactions on Instrumentation and Measurement},
  volume={72},
  pages={1--12},
  year={2023},
  publisher={IEEE}
}

@article{vaswani2017attention,
  title={Attention is all you need},
  author={Vaswani, Ashish and Shazeer, Noam and Parmar, Niki and Uszkoreit, Jakob and Jones, Llion and Gomez, Aidan N and Kaiser, {\L}ukasz and Polosukhin, Illia},
  journal={Advances in neural information processing systems},
  volume={30},
  year={2017}
}

@article{gulrajani2017improved,
  title={Improved training of wasserstein gans},
  author={Gulrajani, Ishaan and Ahmed, Faruk and Arjovsky, Martin and Dumoulin, Vincent and Courville, Aaron C},
  journal={Advances in neural information processing systems},
  volume={30},
  year={2017}
}

@article{kingma2014adam,
  title={Adam: A method for stochastic optimization},
  author={Kingma, Diederik P},
  journal={arXiv preprint arXiv:1412.6980},
  year={2014}
}

@article{youden1950index,
  title={Index for rating diagnostic tests},
  author={Youden, William J},
  journal={Cancer},
  volume={3},
  number={1},
  pages={32--35},
  year={1950},
  publisher={Wiley Online Library}
}

@article{miyato2018spectral,
  title={Spectral normalization for generative adversarial networks},
  author={Miyato, Takeru and Kataoka, Toshiki and Koyama, Masanori and Yoshida, Yuichi},
  journal={arXiv preprint arXiv:1802.05957},
  year={2018}
}

@article{hussain2024realistic,
  title={Realistic labelled pmu data for cyber-power anomaly detection using real-time synchrophasor testbed},
  author={Hussain, Mohammed Mustafa and Sivaramakrishnan, Vasavi and Krishnan, Vignesh and Srivastava, Anurag},
  journal={IEEE Dataport},
  year={2024}
}

@article{yuan2023data,
  title={A data-driven framework for power system event type identification via safe semi-supervised techniques},
  author={Yuan, Yuxuan and Wang, Yanchao and Wang, Zhaoyu},
  journal={IEEE transactions on power systems},
  volume={39},
  number={1},
  pages={1460--1471},
  year={2023},
  publisher={IEEE}
}

@article{abdi2010principal,
  title={Principal component analysis},
  author={Abdi, Herv{\'e} and Williams, Lynne J},
  journal={Wiley interdisciplinary reviews: computational statistics},
  volume={2},
  number={4},
  pages={433--459},
  year={2010},
  publisher={Wiley Online Library}
}

@article{malhotra2016lstm,
  title={LSTM-based encoder-decoder for multi-sensor anomaly detection},
  author={Malhotra, Pankaj and Ramakrishnan, Anusha and Anand, Gaurangi and Vig, Lovekesh and Agarwal, Puneet and Shroff, Gautam},
  journal={arXiv preprint arXiv:1607.00148},
  year={2016}
}

@article{gornitz2013toward,
  title={Toward supervised anomaly detection},
  author={G{\"o}rnitz, Nico and Kloft, Marius and Rieck, Konrad and Brefeld, Ulf},
  journal={Journal of Artificial Intelligence Research},
  volume={46},
  pages={235--262},
  year={2013}
}

@inproceedings{liu2008isolation,
  title={Isolation forest},
  author={Liu, Fei Tony and Ting, Kai Ming and Zhou, Zhi-Hua},
  booktitle={2008 eighth ieee international conference on data mining},
  pages={413--422},
  year={2008},
  organization={IEEE}
}

@article{zheng2021generative,
  title={Generative adversarial networks-based synthetic PMU data creation for improved event classification},
  author={Zheng, Xiangtian and Wang, Bin and Kalathil, Dileep and Xie, Le},
  journal={IEEE Open Access Journal of Power and Energy},
  volume={8},
  pages={68--76},
  year={2021},
  publisher={IEEE}
}

@inproceedings{gao2025semi,
  title={Semi-Supervised Anomaly Detection through Denoising-Aware Contrastive Distance Learning},
  author={Gao, Jianling and Tao, Chongyang and Sun, Zhenchao and Jiang, Xiya and Ma, Shuai},
  booktitle={Proceedings of the ACM on Web Conference 2025},
  pages={2111--2119},
  year={2025}
}

@article{wang2020linformer,
  title={Linformer: Self-attention with linear complexity},
  author={Wang, Sinong and Li, Belinda Z and Khabsa, Madian and Fang, Han and Ma, Hao},
  journal={arXiv preprint arXiv:2006.04768},
  year={2020}
}

@article{choromanski2020rethinking,
  title={Rethinking attention with performers},
  author={Choromanski, Krzysztof and Likhosherstov, Valerii and Dohan, David and Song, Xingyou and Gane, Andreea and Sarlos, Tamas and Hawkins, Peter and Davis, Jared and Mohiuddin, Afroz and Kaiser, Lukasz and others},
  journal={arXiv preprint arXiv:2009.14794},
  year={2020}
}

@article{liu2020anomaly,
  title={Anomaly detection for condition monitoring data using auxiliary feature vector and density-based clustering},
  author={Liu, Hang and Wang, Youyuan and Chen, WeiGen},
  journal={IET Generation, Transmission \& Distribution},
  volume={14},
  number={1},
  pages={108--118},
  year={2020},
  publisher={Wiley Online Library}
}

@inproceedings{liu2022swin,
  title={Swin transformer v2: Scaling up capacity and resolution},
  author={Liu, Ze and Hu, Han and Lin, Yutong and Yao, Zhuliang and Xie, Zhenda and Wei, Yixuan and Ning, Jia and Cao, Yue and Zhang, Zheng and Dong, Li and others},
  booktitle={Proceedings of the IEEE/CVF conference on computer vision and pattern recognition},
  pages={12009--12019},
  year={2022}
}

@article{radford2015unsupervised,
  title={Unsupervised representation learning with deep convolutional generative adversarial networks},
  author={Radford, Alec and Metz, Luke and Chintala, Soumith},
  journal={arXiv preprint arXiv:1511.06434},
  year={2015}
}

@article{park2025unsupervised,
  title={Unsupervised Machine Learning Methods for Anomaly Detection in Network Packets},
  author={Park, Hyoseong and Shin, Dongil and Park, Chulgyun and Jang, Jisoo and Shin, Dongkyoo},
  journal={Electronics},
  volume={14},
  number={14},
  pages={2779},
  year={2025},
  publisher={MDPI}
}

@article{xu2022tgan,
  title={TGAN-AD: Transformer-based GAN for anomaly detection of time series data},
  author={Xu, Liyan and Xu, Kang and Qin, Yinchuan and Li, Yixuan and Huang, Xingting and Lin, Zhicheng and Ye, Ning and Ji, Xuechun},
  journal={Applied Sciences},
  volume={12},
  number={16},
  pages={8085},
  year={2022},
  publisher={MDPI}
}

\end{document}